\documentclass[letterpaper, 10 pt, conference]{ieeeconf}  

\IEEEoverridecommandlockouts                              

\usepackage{graphicx}
\usepackage{mathtools}

\usepackage{float} 
\usepackage{amsmath} 

\DeclareMathOperator*{\argmin}{arg\,min}

\usepackage{bbold}
\usepackage{bbm}

\usepackage{algorithm}
\usepackage[noend]{algpseudocode}

\usepackage{hyperref}
\usepackage{textcomp}
\usepackage{dblfloatfix}
\usepackage{booktabs}
\let\labelindent\relax
\usepackage{enumitem}

\usepackage{wasysym}

\title{\LARGE \bf
Registering Articulated Objects With Human-in-the-loop Corrections
}

\author{Michael Hagenow$^{1}$, Emmanuel Senft$^{2}$, Evan Laske$^{3}$, Kimberly Hambuchen$^{3}$, Terrence Fong$^{4}$\\ Robert Radwin$^{5}$, Michael Gleicher$^{2}$, Bilge Mutlu$^{2}$, and Michael Zinn$^{1}$ 
\thanks{This work was supported in part by a NASA University Leadership Initiative (ULI) grant awarded to the UW-Madison and The Boeing Company (Cooperative Agreement \# 80NSSC19M0124).}
\thanks{$^{1}$Michael Hagenow and Michael Zinn are with the Department of Mechanical
Engineering, University of Wisconsin--Madison, Madison 53706, USA
        {\tt\small [mhagenow|mzinn]@wisc.edu}}%
\thanks{$^{2}$Emmanuel Senft, Michael Gleicher, and Bilge Mutlu are with the Department of Computer
Sciences, University of Wisconsin--Madison, Madison 53706, USA
        {\tt\small [esenft|gleicher|bilge]@cs.wisc.edu}}
\thanks{$^{3}$Evan Laske and Kimberly Hambuchen are with NASA Johnson Space Center, 2101 NASA Pkwy, Houston, TX 77058, USA {\tt\small [evan.laske|kimberly.a.hambuchen]@nasa.gov}}
\thanks{$^{4}$Terrence Fong is with the Intelligent Robotics Group, NASA Ames Research Center, Mountain View, CA 94035, USA {\tt\small terrence.w.fong@nasa.gov}}%
\thanks{$^{5}$Robert Radwin is with the Department of Industrial and Systems Engineering, University of Wisconsin--Madison, Madison 53706, USA
        {\tt\small rradwin@wisc.edu}}%
}

\begin{document}

\maketitle
\thispagestyle{empty}
\pagestyle{empty}

\begin{abstract}
Remotely programming robots to execute tasks often relies on registering objects of interest in the robot’s environment. Frequently, these tasks involve articulating objects such as opening or closing a valve. However, existing human-in-the-loop methods for registering objects do not consider articulations and the corresponding impact to the geometry of the object, which can cause the methods to fail. In this work, we present an approach where the registration system attempts to automatically determine the object model, pose, and articulation for user-selected points using nonlinear fitting and the iterative closest point algorithm. When the fitting is incorrect, the operator can iteratively intervene with corrections after which the system will refit the object. We present an implementation of our fitting procedure for one degree-of-freedom (DOF) objects with revolute joints and evaluate it with a user study that shows that it can improve user performance, in measures of time on task and task load, ease of use, and usefulness compared to a manual registration approach. We also present a situated example that integrates our method into an end-to-end system for articulating a remote valve.
\end{abstract}
\section{INTRODUCTION}\label{sec:intro}
Robotic solutions can enable tasks to be completed remotely in inaccessible environments such as space operations. To ensure success in these mission-critical scenarios, task plans are typically specified remotely where an operator programs the robot actions \cite{fong2013space}. Many of these actions involve manipulating objects in the remote environment, such as valves and switches, that often include some form of articulation (i.e., joints). When programming interactions, ideally the operator would be able to accurately and easily register the pose and state (e.g., articulation) of objects from visual information of the scene. Unfortunately, existing automated recognition methods cannot reliably identify articulated objects, and manual registration methods are burdensome for end users. In this work, we propose a human-in-the-loop method for registering articulated objects that reduces user input through semi-automated fitting and maintains accuracy by allowing operator verification and corrections.

\begin{figure}
\centering
\includegraphics[width=3.30in]{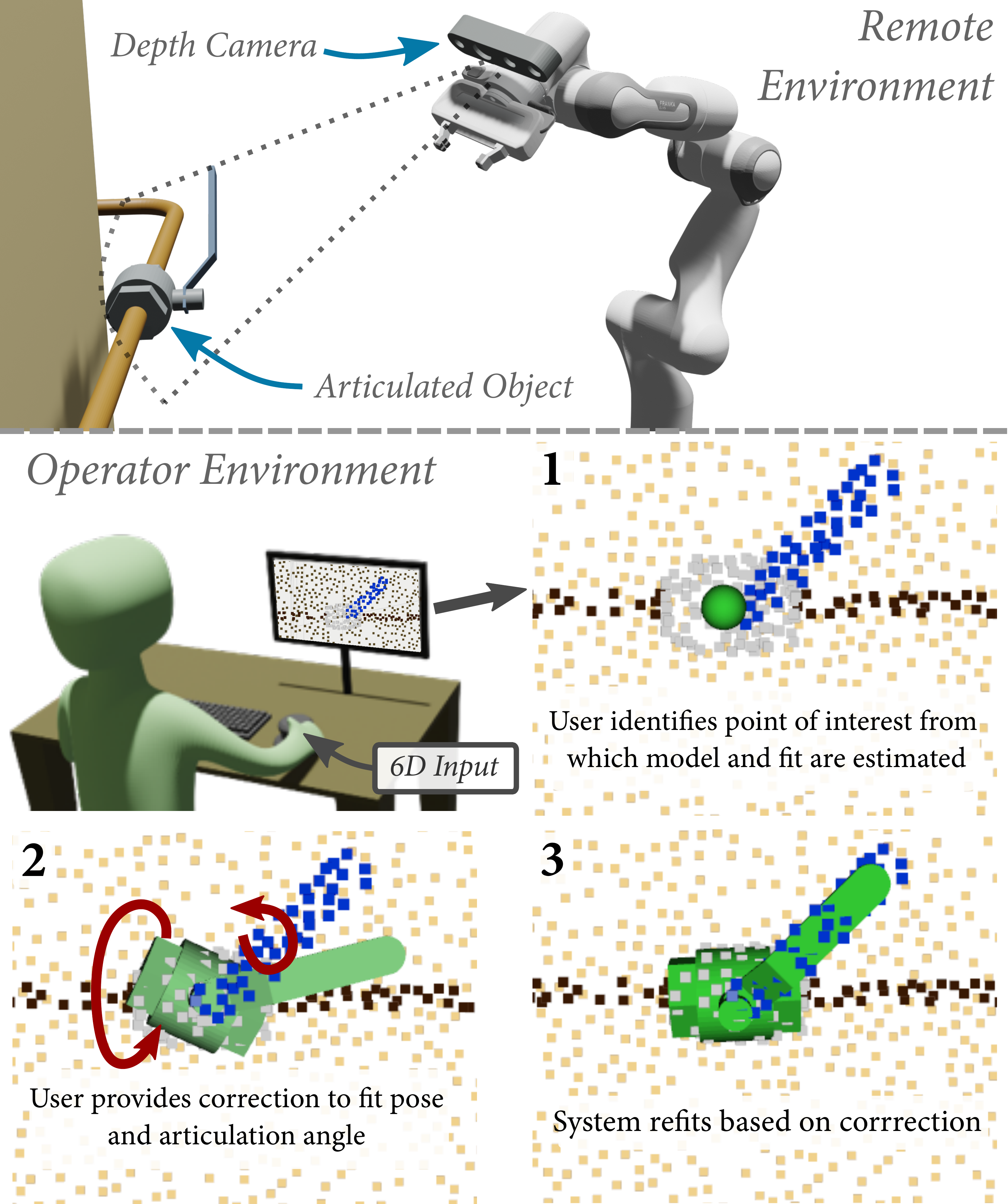}
\caption{Prototypical application for the proposed system. (Top) A depth camera on a robot provides a localized point cloud including a ball valve. (Bottom) (1) The user selects a point on the point cloud and the system automatically fits the model, pose, and articulation. (2) The user provides a correction to the fit pose and articulation (3) after which the system refits. The final articulated model fit can be used for task planning.}
\label{fig:teaser}
\vspace{-15pt}
\end{figure}

Accurately registering objects is a preliminary step for many methods for planning remote robot manipulations. Specifically, knowledge of the pose and object state allows for robots to plan appropriate trajectories and to set appropriate task and compliance frames for physical interaction. For example, Affordance Templates (ATs) \cite{hart2014affordance}\cite{HartAffordance2015} are a method to manipulate known objects under constraints such as object joint limits or maximum applied torques. However, the current approach to register objects in the Affordance Template framework relies on manually positioning and aligning a set of interactive markers to identify the object pose and further manually parameterizing the object state (e.g., valve angle). Semi-automated fitting approaches, summarized in Section \ref{sec:relatedwork}, have been proposed to further automate registration by estimating the pose of known object meshes, however, these approaches do not explicitly consider articulation. In many methods, the object is registered through a static mesh and thus mismatches between the articulation of the actual object and the articulation of the mesh (e.g., the real valve is closed and valve model is open) may cause these methods to improperly fit objects.

In this work, we describe a \textit{human-in-the-loop} method for registering both articulated and non-articulated instances of known object models. Our method is designed to provide a high level of automation where the model, pose, and articulation angle are automatically estimated for user-identified objects of interest and a remote operator provides corrections only when desired. Our fitting method uses nonlinear fitting and the iterative closest point algorithm, which is commonly used in other human-in-the-loop methods, to enable simultaneous fitting of articulations alongside object poses. Our core contributions are (1) describing a mostly automated human-in-the-loop workflow for registering revolute 1-DOF objects, (2) providing an open-source implementation of our method where users can register articulated models using a graphical interface and provide corrections using a 6D input (see Figure \ref{fig:teaser}), and (3) conducting a user-study-based evaluation showing that our method improves perceived usefulness and ease of use as well as reduces workload and the time required to accurately register object instances compared to existing manual specification techniques. We also provide a situated example that integrates our method as part of an end-to-end object manipulation pipeline.

\section{RELATED WORK} \label{sec:relatedwork}
Our work combines nonlinear fitting of known articulated models with human-in-the-loop corrections. To contextualize our contributions, we provide a brief review of methods for estimating articulated models and methods for semi-automated object registration.

Existing methods for registering articulated objects focus on automatic recognition of known objects or inference of permissible articulation for novel objects. Huang et al. \cite{huang2013detecting} explore registration of similar object classes (e.g., valves), but focus on automatic segmentation using RANSAC and do not consider the articulation of the object models. Many methods focus on estimating the pose of known highly articulated models, such as characters or skeletons \cite{prajapati2013fast, chang2008automatic}. As an example, Anguelov et al. \cite{anguelov2012recovering} decompose an articulated mesh into approximate rigid parts and use Expectation Maximization (EM) to estimate part assignments and transformations. Other recent methods focus on estimating articulation of novel objects though images \cite{mo2021where2act,jain2021screwnet,zeng2020visual, desingh2019factored}, physical interaction \cite{subramani2020method, sturm2011probabilistic}, or combining physical interaction and vision data \cite{martin2019coupled}. For example, Jain et al. \cite{jain2022distributional} learn a distribution over articulation model parameters for novel objects with different degrees of freedom. Generally, the results of these automated methods are impressive, however, the classification accuracies are insufficient for critical applications (e.g., average of 20 degrees error for the articulation axis in \cite{jain2022distributional}). Thus, there is still a need for human-in-the-loop approaches for validation and corrections in critical applications.

Previous work has looked at human-in-the-loop techniques to identify objects for manipulation, but has not considered object articulation as part of a semi-automated fitting process. Butler et al. \cite{butler2017interactive} propose and evaluate an interactive segmentation method where candidate objects are identified using a 2D selection interface on a point cloud. Masnadi et al. \cite{masnadi2019sketching}\cite{masnadi2020affordit} propose methods for affordance specification by sketching relevant articulations on a touchscreen interface, however, the method is manual and does not use knowledge of the environment to refine or assist the operator in registering affordances. Kent et al. \cite{kent2020leveraging} propose a method for human-in-the-loop object grasp specification based on user selection from a small list of candidate grasp poses. Jorgensen et al. \cite{jorgensen2019deploying} use Iterative Closest Points (ICP) to automatically register static meshes based on rough positioning of object models. Finally, Ye et al. \cite{ye2021human} propose a snap-to-grid approach using Monte Carlo localization where static objects are first roughly positioned using interactive markers, followed by automated system pose refinement.

Our method combines a human-in-the-loop workflow with existing technical methods for fitting articulated objects. Our work is contextualized in visual identification of articulated objects from a point cloud for instances where exploratory motion and physical interaction cannot be used to identify object properties (e.g., gas leak or inspection prior to manipulation). By automatically estimating the model, pose, and articulation of an object at a user-selected point, our method operates at a higher level of automation than many existing semi-automated approaches (i.e., additional user input is only required when the system incorrectly identifies or improperly registers the object at the user-selected point). Simultaneously, by maintaining a human in the loop, the operator remains in control of the system and can verify and provide corrections to the system output to assure correctness for mission-critical operations.
\section{METHOD}
In our method, operators provide corrections as part of a semi-automated fitting method for articulated models. In this section, we describe our interaction design and fitting approach for both articulated and non-articulated objects.

\subsection{Operator Interaction Framework}
Our framework was designed to minimize the amount of input required from the operator during object registration. The operator provides corrections via a remote system that consists of an environment visualization (e.g., monitor, VR) and an input for providing corrections (e.g., 6D input, VR controller). The system is instantiated with a scene (e.g., point cloud capture) and a set of candidate models that describe the objects' geometry including permissible articulation. Objects are registered in the environment as follows:

\begin{enumerate}[leftmargin=*]
  \item The operator selects a point on the object in the 3D scene. This initialization bounds the search space and circumvents the need for scene segmentation.
  \item The pose and articulation of candidate models are fit using the approach in Section \ref{sec:fitting}. For each model ($m$), a likelihood is constructed from the residual ($e_m$) and number of sampled mesh points ($n_m$): $L(m)=n_m/e_m$.
  \item The mostly-likely fit is displayed to the user. If the model is incorrect, the user can cycle to the next most likely fit. If the pose or articulation is not correct, the user provides relative corrective input. After the operator finishes providing input (i.e., a brief pause), the system can refit the object. This process of providing corrections and system refitting can be performed iteratively.
  \item The operator can repeat steps $2\--4$ as necessary to register additional objects within the scene.
\end{enumerate}

\subsection{Fitting Procedure} \label{sec:fitting}
Our model-fitting approach is based on the iterative closest point (ICP) algorithm \cite{besl1992method}. While other registration algorithms would also be acceptable (e.g., particle filters \cite{ye2021human}), we chose ICP based on its speed, prevalence, and convergence when initialized near the correct pose. Though it requires appropriate mesh sampling and is susceptible to local minima, we believe it is a strong candidate for our method where an operator's corrections will typically adjust the object pose and articulations to be close to the globally optimal fit when the initial fitting is incorrect. In the following subsections, we describe the general ICP algorithm, the method for solving the ICP problem for non-articulated models, and finally the modified optimization and approach that is used to solve articulated models.

\subsubsection{ICP Algorithm}
Given a scene, a user-selected point, and a set of models, we desire to fit each candidate model to the scene data. The ICP algorithm calculates the transformation that optimally aligns two sets of corresponding points. We calculate this transformation using the point cloud scene and the sampled candidate model mesh as the two sets of points. The correspondences are determined using a data structure that enables efficient nearest-neighbor checks (e.g, k-d tree). The ICP optimization is described by: 
\begin{equation}\label{eq:icpbase}
\begin{aligned}
    \argmin_{\textbf{R},\textbf{t}} \quad & \sum\limits_{i} w_{i}\lVert\textbf{R}\textbf{p}_{i}+\textbf{t}-\textbf{s}_{i}\rVert^{2}\\
    \textrm{s.t.} \quad & \textbf{R}\in \mathit{SO}(3)
\end{aligned}
\end{equation}
where $\textbf{R}$ and $\textbf{t}$ are the optimal rotation and translation, $i$ is the point index, $w$ is a weighting factor, $\textbf{p}_{i}$ is the source point (e.g., mesh), and $\textbf{s}_{i}$ is the closest scene point.  

\subsubsection{Fitting of Non-articulated Models}
When a model does not include articulation, we solve for the optimal pose using the ICP optimization in Equation \ref{eq:icpbase}. This optimization can be solved in closed form and computed efficiently using the Singular Value Decomposition (SVD) \cite{eggert1997estimating}. To provide improved robustness to occlusions and partial
views, we reject outlier samples where the distance between corresponding points in the two sets, $d_{i}$, exceeds the model’s maximum width and also weigh the correspondences based on the distance.
\begin{equation}
    w_{i} = (1+d_{i})^{-1}
\end{equation}

The result of ICP is subject to initialization and the algorithm often converges to local minima because of the non-convexity of the problem. To reduce this occurrence, we use a random restarts procedure. For each restart, the initial conditions are drawn from random orientations and translation offsets. The offset from the initial user-selected point is drawn from a normal distribution that is bounded by the model's maximum width (i.e., model diameter).

\subsubsection{Fitting of Articulated Models}
When a model includes articulation, we must solve for the value of the articulation (e.g., joint angle) in addition to the base rotation and translation during the alignment step. We focus on the common case of one articulation (e.g., valves). Thus, our model consists of two meshes where each mesh corresponds to one link of the articulated model and requires a modified optimization:
\begin{equation}
\begin{aligned}
    \argmin_{\textbf{q},\textbf{t},\theta} \quad & \sum\limits_{i} w_{i}\lVert\textbf{FK}(\textbf{p}_{i},\textbf{q},\textbf{t},\theta)-\textbf{s}_{i}\rVert^{2}\\
    \textrm{s.t.} \quad & \lVert\textbf{q}\rVert = 1, 
     \theta_{min} \leq \theta \leq \theta_{max}
\end{aligned}
\end{equation}
where \textbf{q} is the optimal rotation expressed as a quaternion (with a corresponding normalization constraint), $\theta$ is the articulation parameter (e.g., joint angle), $\theta_{min}$ and $\theta_{max}$ are the joint limits, and $\textbf{FK}$ is a forward kinematics function that determines the location of a mesh point after applying the appropriate chain of rotations and translations to each of the mesh links. Unlike the fitting for non-articulated objects, this optimization cannot be solved in closed form and requires a constrained nonlinear optimization solver.

One other challenge in fitting the articulated models is that the nonlinear approach is less performant than the linear SVD approach. As a consequence, running random restarts on the nonlinear fitting notably increases the fitting time which can degrade the user experience.

To address this, we employ a hybrid approach. For each of the random restarts, we assign a fixed random articulation angle and solve for the object pose using the SVD method. Once all restarts have completed, we run the nonlinear fitting once where the initial conditions are taken from the random restart with the lowest residual (i.e., the best fit). Empirically, this hybrid approach increases performance and produces similar results to running nonlinear fitting for each restart.

\begin{figure}
\centering
\vspace{5pt}
\includegraphics[width=3.1in]{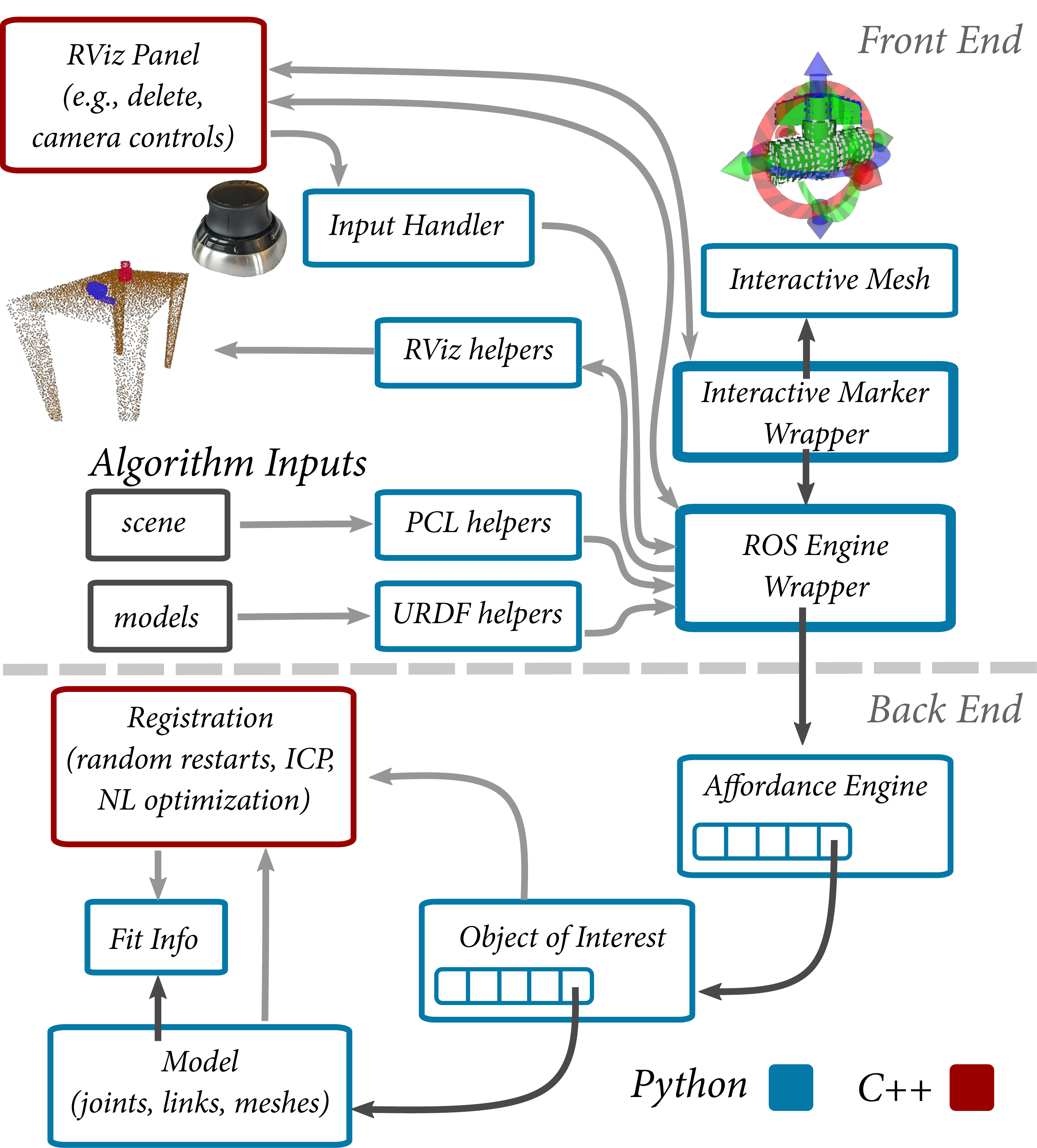}
\caption{Implementation class diagram for human-in-the-loop corrections of articulated objects. The implementation is split into a front end focusing on input and visualization tools in ROS and a back end that stores objects of interest and fits models. Light gray arrows imply data flow. Dark gray arrows originating from inside classes imply encapsulation. Either the \emph{ROS Engine Wrapper} (SpaceMouse) or \emph{Interative Marker Wrapper} represents the main class depending on input method.}
\label{fig:schematic}
\vspace{-15pt}
\end{figure}

\section{IMPLEMENTATION}
We developed an open-source implementation\footnote{\url{https://gitlab.com/mhagenow/HIL_Obj_Registration}} of our method. In the implementation, we refer to fitting object articulations as affordances. The implementation consists of a Python/C++ back end and ROS \cite{quigley2009ros} front end as shown in Figure \ref{fig:schematic}. This section describes the main elements such as the data engine class, inputs/visualization, and registration.

\subsection{Data Engine}
All tracking is maintained by a back-end class, called the \emph{Affordance Engine} (AE), designed with an API for flexible use by various inputs and outputs. The AE is responsible for storing the environment point cloud and candidate object models, capturing operator input (e.g., new search points, deletions, corrections), storing fits, and providing state information for visualization. The AE maintains a list of user-selected search points. Each of these points contains fitting information for each of the available models. Maintaining all model fit information allows for cycling between models when the system incorrectly identifies the model.

The class is instantiated with a point cloud scene and a list of candidate models. The scene can be loaded from a PCD file or ROS PointCloud2 message. The models can either be specified as an STL file (for stationary objects) or as a Unified Robot Description Format (URDF) for articulated objects. For the fitting, the URDF is required to specify STLs for each of the links (e.g., valve base, valve handle). The current system only supports 1-DOF revolute joints.

\subsection{Input and Visualization}
\begin{figure}
\centering
\includegraphics[width=3.30in]{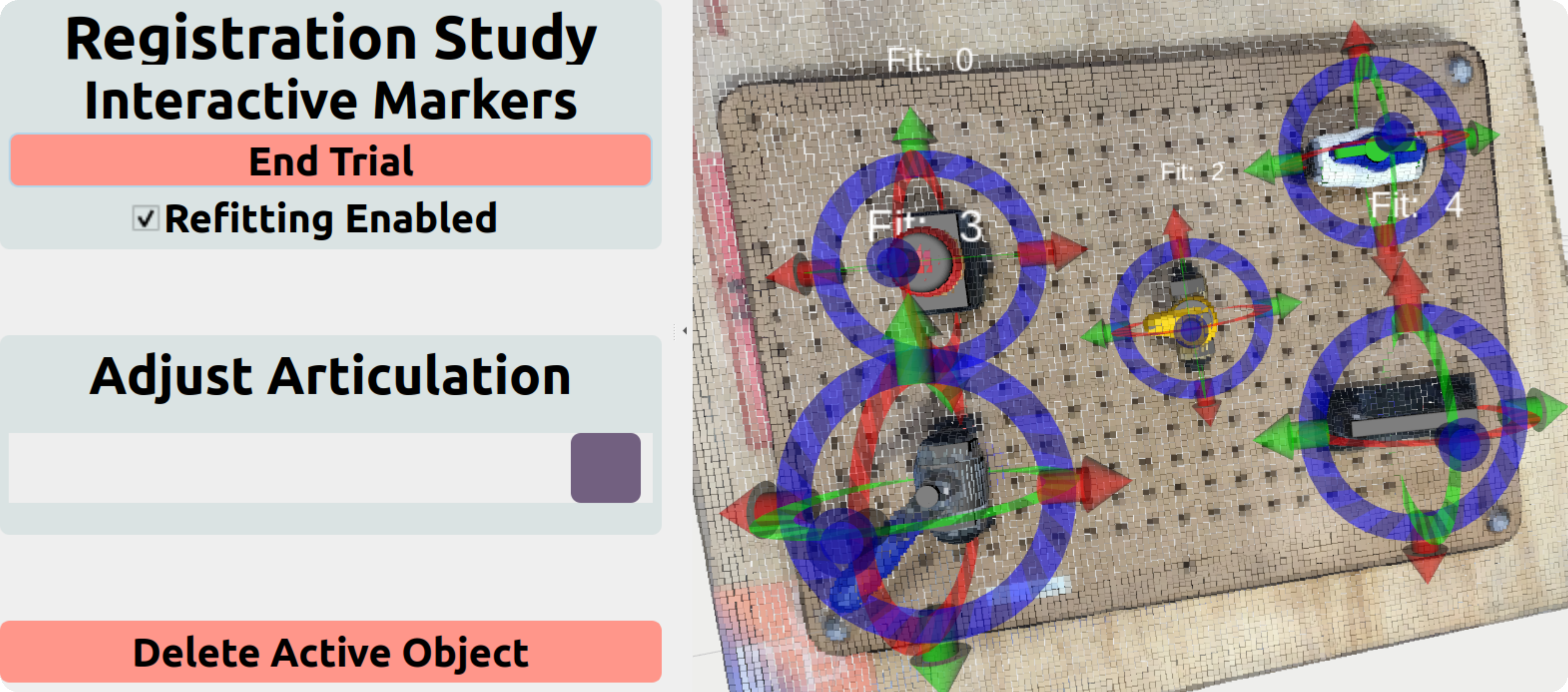}
\caption{Example of the user-study interface for the semi-automated interactive marker fitting. The text of the interface has been enlarged for readability. The interface contains a start/stop trial button, a delete button, a toggle for refitting, and a slider to articulate the valves. The manual methods used similar interfaces without the refitting checkbox. The SpaceMouse interface was similar, but does not have the 'Adjust Articulation' slider. }
\label{fig:expconditions}
\vspace{-15pt}
\end{figure}

Our implementation leverages RViz for operator input and to visualize registered objects in the scene. Users interact with the point cloud using the RViz `publish point' feature. If there isn't a current object of interest at the selected point, the system will fit all models and display the mesh of the best-fit model. If there is an existing fit, the `publish point' action will make the object active or cycle to the next most likely model if the object is already active. Users are able to delete object fits and provide additional input (e.g., an articulation slider for some methods) using a custom RViz panel on the left side of the screen, as shown in Figure \ref{fig:expconditions}.

The implementation focuses on two input methods: a 6D SpaceMouse\textregistered\footnote{\emph{ \label{trademark}3Dconnexion. Trade names and trademarks are used in this report for identification only. Their usage does not constitute an official endorsement, either expressed or implied, by the National Aeronautics and Space Administration.}} and 6D interactive markers \cite{gossow2011interactive}. The SpaceMouse was selected to test a 6D relative input for the corrections. The interactive marker approach was chosen based on its prevalence in existing systems.

When using the SpaceMouse, corrections to the pose are applied based on the 6D input twist. Operators are also able to provide corrections to the articulation using two buttons on the side of the mouse. The twist is applied relative to the current viewpoint of the camera in Rviz. The 6D input from the SpaceMouse is scaled exponentially (i.e.,  to calculate the final twist applied to the model. This scaling allows for finer resolution inputs when applying small twists while maintaining the ability to apply large twists for coarse corrections (i.e., the model fit is far from the correct pose).

When using interactive markers, corrections to pose are made using the six visual interactive marker controls. To avoid further cluttering the visual space surrounding the object, corrections to the articulation angle are made using a slider in the RViz panel on the left side of the screen. Operators also have a second option to cycle models by right clicking on the interactive markers which displays a menu of available models sorted by the likelihood.

\subsection{Registration}
To increase performance and improve the operator experience, the registration fitting algorithm was developed in C++. Data is passed to the registration class using Cython \cite{behnel2010cython}. Each model is fit to the scene data in parallel using OpenMP \cite{chandra2001parallel}. For each model, the ICP uses 25 random restarts.

Corresponding points between the model and scene are aligned using an optimized k-d tree\footnote{\url{https://github.com/Jaybro/pico_tree}}. For performance, each model was sampled uniformly using 400 points and the point cloud scene was voxelized to 5 mm and culled to a sphere around the user-selected point with a radius bounded by the maximum width of the candidate model. The number of points was empirically tuned to balance accuracy and performance. The exit criteria for the ICP algorithm was tuned empirically to 100 iterations or when the per-point residual is changing by less than $2.5\times10^{-8}$ m. The linear SVD optimization leverages Eigen \cite{eigenweb} and the nonlinear optimization uses COBYLA\cite{powell1998direct}. After the operator has provided corrections, the system refits the object using the corrected pose and articulation as the starting point. This refitting process uses the same ICP routine as the initial fitting, but without the random restarts.


\section{EXPERIMENTAL EVALUATION}

\begin{figure}
\centering
\includegraphics[width=3.10in]{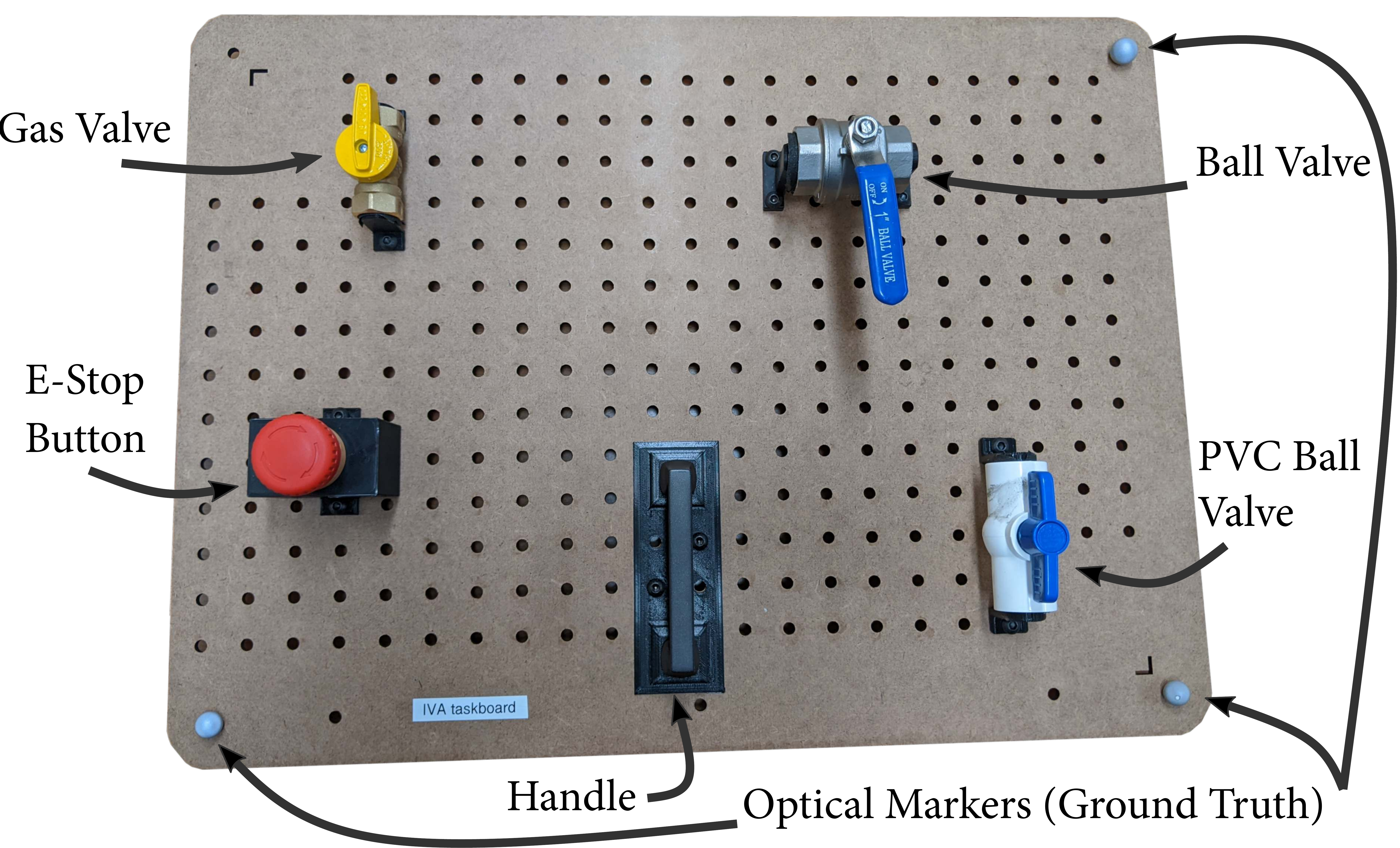}
\vspace{-5pt}
\caption{Taskboard consisting of optical markers for registering the ground truth and an M6 grid of holes for mounting. Experimental objects consist of three objects modeled with articulations: gas valve, ball valve, and PVC ball valve as well as two objects without articulation: e-stop button and handle.}
\label{fig:taskboard}
\vspace{-15pt}
\end{figure}

To demonstrate our proposed method, we conducted a preliminary user study and designed a situated example of an end-to-end system to remotely manipulate objects. The purpose of the user study was to understand choices in input method and to evaluate the performance and usability of our method compared to existing methods. The situated example was a demonstration of integrating our method into an end-to-end system to manipulate objects.

\subsection{User Evaluation of Human-in-the-loop Fitting}\textbf{}
Our study involved 15 participants (9M/6F), aged 18--22 ($M=19.6$, $SD=1.2$), recruited from the University of Wisconsin--Madison campus. The procedure was administered under a protocol approved by the Institutional Review Board
(IRB) of UW--Madison. The study followed a $2\times2$ within-subjects design and was designed to last up to ninety minutes. The independent variables included whether or not the fitting/refitting was enabled and the input device (i.e., SpaceMouse, interactive markers). The hybrid fitting was used for objects including articulation. When the fitting was disabled, clicking a point in the scene instantiated the first model in the chosen position with the identity matrix as its rotation and zero articulation.

\subsubsection{Procedure}
Following informed consent, participants were briefed on the structure of the experiment and shown a six-minute video that explained how to register objects under each of the experimental conditions. After watching the video, participants practiced for up to ten minutes with each of the conditions on an example taskboard. To gain familiarity with the input devices, participants practiced with the manual condition of each input method before the semi-automated fitting for the same input method. The order of the input devices was counterbalanced. Participants were allowed to ask questions about the system during training. There were no intentional errors introduced during the registration. Participants were instructed that they did not need to provide corrections if the initial fit was accurate.

During the experiment, participants registered a different taskboard than the training. To limit experiment length and participant fatigue, participants had up to ten minutes to complete each trial. The order of the conditions followed a balanced Latin square design where no order of conditions was repeated. In each trial, the orientation of the board was randomly rotated (up to 0.3 radians per axis). The participant clicked a button to signal the start and end of each trial.


\begin{figure*}[b]
\centering
\includegraphics[width=.95\textwidth]{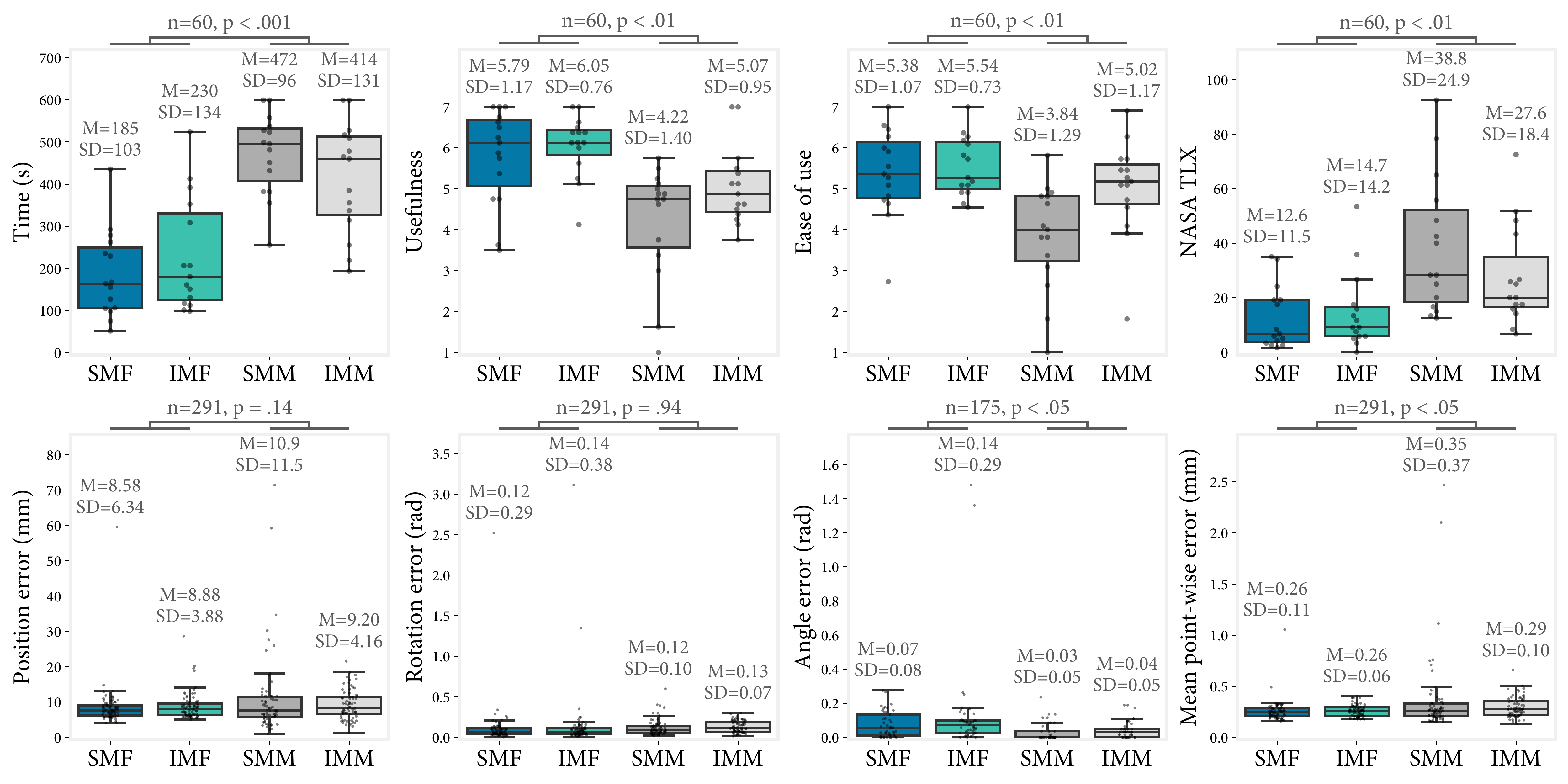}
\vspace{-5pt}
\caption{Individual data overlaid on a Boxplot for each study condition. The conditions are SpaceMouse with fitting (SMF), interactive marker with fitting (IMF), SpaceMouse manual (SMM) and interactive marker manual (IMM). Significance values are included for the effect of fitting.}
\label{fig:results}
\end{figure*}

\subsubsection{Hypotheses}
Our study consisted of two independent variables: whether the semi-automated fitting was enabled and the input method. Our investigation of the semi-automated fitting (i.e., the fitting condition) was designed to confirm the benefits of our proposed method whereas our investigation of the input method was exploratory. For the fitting, we evaluated three hypotheses as part of our study: 
\begin{itemize}[leftmargin=*]
  \item \textbf{H1}: Fitting would result in a lower time on task compared to manual registration.
  
  \item \textbf{H2}: Fitting would be rated as easier to use, more useful, and having a lower taskload compared to manual.
  
  
   \item \textbf{H3}: No significant differences in performance would be observed between fitting and manual registration.
   
  
\end{itemize}

\subsubsection{Apparatus}
Participants were seated and used a keyboard, mouse, and SpaceMouse (when applicable) to provide input. Each trial consisted of fitting five objects that were mounted to a high-density fiberboard taskboard as seen in Figure \ref{fig:taskboard}. The five objects consisted of three objects that were fit with articulations (\emph{ball valve}, \emph{PVC ball valve}, and \emph{gas valve}) and two objects fit without articulations (\emph{handle} and \emph{e-stop}) that were represented by a static mesh. While the \emph{e-stop} button does articulate, the button throw was too short for identification from visual information and thus, it was treated as static. The five objects were selected as representative hardware from intra-vehicular activities on the International Space Station (ISS). To simplify collecting the ground-truth value of the articulation, one valve was completely open, one was half-way open, and one was completely closed (i.e., 0, 45, or 90 degrees). Point clouds were captured using an Occipital Structure sensor. To register the ground truth of the objects, three optical markers were attached to known locations on the task board.

\begin{figure*}[t]
\centering
\includegraphics[width=.95\textwidth]{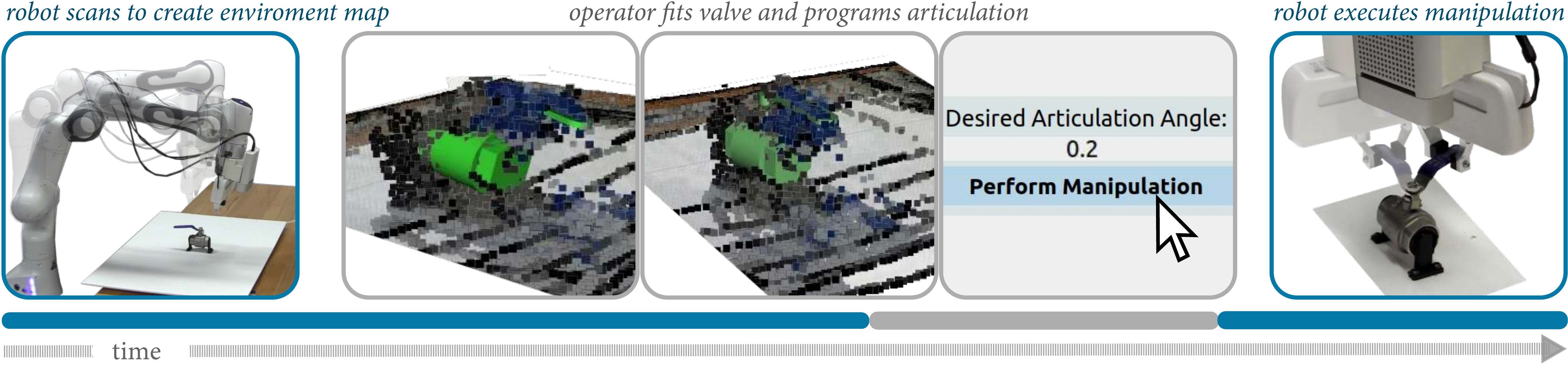}
\vspace{-5pt}
\caption{Situated example timeline (length of bars correspond to relative event length). The robot scans the environment and the user creates a valve fit and then adjusts the pose and articulation angle. The operator programs the valve to rotate to 0.2 radians after which the robot executes the action.}
\label{fig:sitexample}
\vspace{-15pt}
\end{figure*}

\subsubsection{Metrics}
After each condition, participants filled out the NASA Task Load Index (TLX) \cite{NASATLX} and the \emph{Usefulness} and \emph{Ease of Use} sections of the USE Questionnaire \cite{lund2001measuring}. After completing the experiment, participants completed a brief interview where they were asked for feedback and preferences on the fitting and input methods.

We also collected the total time, system time for fitting/refitting, and tracked the final registered objects including the model selected, position, rotation, and articulation angle (when applicable). We compared these fits to the ground truth by selecting the closest objects and calculating differences in position, rotation, and articulation. The error in rotation was calculated from the ground truth quaternion, $\textbf{q}_{gt}$, and registered quaternion, $\textbf{q}_{reg}$:
\begin{equation}
    e_{rot} = 2\arccos(|\operatorname{Re}(\textbf{q}_{gt}^{-1} \star \textbf{q}_{reg})|)
\end{equation}
where $\star$ denotes quaternion multiplication. For certain objects (e.g., PVC ball valve, handle, and e-stop), there were two poses in which the object appearance was identical (e.g., 180 degree rotation around the button axis of the e-stop) and thus, we calculated the error for both poses and chose the smaller of the rotation errors. Additionally, as it was possible the articulated models could have one link (e.g., the valve handle) accurately positioned without having the correct pose or articulation angle, we also reported the mean point-wise difference between the ground truth mesh and registered mesh as another proxy for fit quality. The metrics were analyzed using a two-way repeated measures ANOVA to report both individual and interaction effects (corrected as needed using Greenhouse-Geisser). For each metric, we report mean ($M$), standard deviation ($SD$), observation significance ($p$), and the test critical value ($F$).

\subsubsection{Results}
Figure \ref{fig:results} summarizes the results of the user study. We report the full statistics, including the exploratory condition, in the following paragraphs. 

\textbf{H1} was fully supported by our results. For time on task, we found a trend towards significance for the interaction effect ($F(1,14)\!=\!3.46,p \!=\! .08$), a significant effect for fitting ($p \!<\! .001$), but no significant effect for input method ($p \!=\! .74$).

\textbf{H2} was also fully supported. For usefulness, we found no significant interaction effect between the conditions ($F(1,14)\!=\!1.15,p \!=\! .3$), a significant effect for fitting ($p \!<\! .01$), and a significant effect for input method ($p \!<\! .05$). For ease of use, we found a significant interaction effect between the conditions ($F(1,14)\!=\!5.62,p \!<\! .05$), a significant effect for fitting ($p \!<\! .01$), and a significant effect for input method ($p \!<\! .01$). For the taskload, we found a significant interaction effect between the conditions ($F(1,14)\!=\!5.34,p \!<\! .05$), a significant effect for fitting ($p \!<\! .01$), and a marginally significant effect for input method ($p \!=\! .05$).

\textbf{H3} was partially supported. For position error, we found no significant interaction effect between the conditions ($F(1,14)\!=\!1.65,p \!=\! .22$). We did not find a significant effect for fitting ($p \!=\! .14$) and no significant effect for input method ($p \!=\! .29$). For rotation error, we found no significant interaction effect between the conditions ($F(1,14)\!=\!0.80,p \!=\! .39$). We did not find a significant effect for fitting ($p \!=\! .94$) and no significant effect for input method ($p \!=\! .27$). For the articulation angle, we found no significant interaction effect between the conditions ($F(1,14)\!=\!1.19,p \!=\! .29$). We did find a significant effect for fitting ($p \!<\! .05$) but no significant effect for input method ($p \!=\! .1$).
For the mean point-wise error, we found no significant interaction effect between the conditions ($F(1,14)\!=\!1.51,p \!=\! .24$). We did find a significant effect for fitting ($p \!<\! .05$) but no significant effect for input method ($p \!=\! .19$).

The average fitting and refitting time for the fitting conditions was $0.88\pm 0.12$ seconds and $0.23\pm 0.20$ seconds respectively. All participants preferred the semi-automated fitting over the manual fitting. However, five participants noted preference to disable refitting. Seven participants preferred the SpaceMouse and eight participants preferred the interactive markers. The majority of participants were able to complete tasks without any major errors. Two participants ran out of time on the SpaceMouse manual trial. One participant ran out of time on the interactive marker manual trial. One participant only fit four objects during the SpaceMouse manual trial and another participant only fit four objects during the interactive marker manual trial.

\subsection{Situated Demonstration}
We designed a situated example to illustrate the intended application of our method. As shown in Figure \ref{fig:sitexample}, the context was to program a remote robot to actuate a valve to a specified state (i.e., angle). The setup consisted of a Franka Emika robot that was operated with joint compliance (100 N/m) to facilitate safe environment interaction and provide partial robustness to registration and localization error. The robot was equipped with a gripper and Kinect Azure RGB-D camera mounted to the distal link with a known transform.
The environment was mapped from the Kinect's point cloud using RTAB-Map \cite{labbe2019rtab} where the localized camera position was used as the ground-truth odometry.

The task was to actuate a 1 inch ball valve, which was initially open, to a partially closed configuration (0.2 radians). The robot first scanned the scene using a pre-programmed trajectory within the reachability of the platform. Next, the operator registered the object to the scene and specified the desired angle for the valve. Finally, the robot executed the manipulation to the desired angle using a behavior parameterized with the starting and ending joint angle. After the manipulation, the robot could optionally rescan the environment to confirm the final state. The example is shown in its entirety, including rescanning, in the supplemental video. The task was completed three times and in each trial, the operator provided corrections to the initial valve fit and the system was able to articulate the valve.
\section{DISCUSSION}

Our study validated that our human-in-the-loop method for registering articulated objects reduces time on task and improves usability, ease of use, and task load compared to manual specification (\textbf{H1} and \textbf{H2}). We did, however, find significant differences in performance depending on whether the fitting was enabled (\textbf{H3}). The articulation error was found to be lower with manual conditions, however, this may be attributed to experiment bias where two of the articulations were extreme angles (e.g., 0 and 90 degrees) that were simple to specify manually and also possibly that the achievable angles of the hardware joints did not exactly equal the ground-truth specifications. The mean point-wise error was lower for semi-automated fitting, which may indicate the semi-automated fitting tended to be more precise. However, for both the articulation angle and mean point-wise error, the absolute values of error in every condition were minimal. These results in \textbf{H3} indicate that the fitting assistance may both positively and negatively impact the quality of fits, however, we believe future studies are required to properly isolate and quantify these impacts (e.g., minimizing experiment bias and using a precision system for ground-truth collection).

We also found that not all participants preferred the refitting functionality. In several cases, this was the result of registration strategy, where participants would move object fits away from the board to improve their visual inspection of articulation angle and orientation. This often would conflict with the refitting that would refit the object based on the displaced location. Two participants noted that the ability to toggle the refitting was a useful feature to include.


Participants generally found the SpaceMouse to have a lower ease of use and usability, yet the SpaceMouse did not have a significant decrease in performance (e.g., time, error) compared to the interactive markers. We observed some interaction effects. Participants trended towards spending more time and reported a higher task load when using SpaceMouse manually compared to interactive markers, but trended towards less time and reported a lower task load when using the SpaceMouse with fitting compared to interactive markers. These findings were consistent with post-experiment interviews. Many participants noted that the SpaceMouse was harder to learn, more difficult to provide precise input, and less intuitive compared to interactive markers. However, some participants noted that the SpaceMouse made providing coarse inputs faster, which when coupled with refitting, could allow for accurate fitting in less time.

\emph{Limitations and Future Work--}Our work requires that object models are known a priori and that the desired articulation can be identified from the object appearance. In the future, we plan to explore the impact of object model quality and ways to minimize the effort to create new models (e.g., on the fly from point clouds). Certain objects, such as valves with circular handles or buttons with short throws, cannot be reliably registered from visual information. We plan to supplement our system with a confidence metric for the articulation angle that is based on whether the articulation can be identified visually. In this work, we leveraged ICP for fitting. In the future, we would like to explore the impact of using other technical tools (e.g., particle filters). 

In our user study, participants only interacted with high-quality point clouds provided by the Occipital Structure sensor. In the future, we would like to improve our method for lower quality point clouds, such as those in the situated demonstration. The evaluation only assessed spaced-out objects. Additionally, we plan to study the system in contexts such as space operations and general cluttered manipulation environments. We also plan to further assess the impact of other practical details such as varied noise conditions or the impact of particular object geometries on fitting. Our preliminary study involved only fifteen participants. In the future, we plan to conduct larger-scale studies. In this work, we only assessed single revolute joints. We would like to extend our method to more highly-articulated models and include additional classes of articulation, such as prismatic joints. The situated demonstration was designed as a first example to demonstrate the plausibility of including our method in an end-to-end system. Looking forward, we would like to perform more demonstrations and conduct quantitative analysis in a system such as the NASA Valkyrie.

\section{ACKNOWLEDGEMENTS}
We would like to thank Kevin Macauley and Nicole Gathman for assistance in implementation and the NASA Johnson Space Center ER4 Robotic Systems Technology Branch for feedback on the early system, in particular Philip Strawser, Andrew Sharp, Lewis Hill, and Steven Jorgensen.

\bibliographystyle{IEEEtran}
\bibliography{references}

\end{document}